\pdfoutput=1

\documentclass[11pt]{article}

\usepackage[]{acl}

\usepackage{times}
\usepackage{latexsym}

\usepackage[T1]{fontenc}

\usepackage[utf8]{inputenc}

\usepackage{microtype}

\usepackage{multirow}
\usepackage{adjustbox}
\usepackage{hyperref}
\usepackage{comment}
\usepackage{amsmath}

\newcommand\todo[1]{{\textcolor{red}{#1}}}

\title{Injecting Wiktionary to improve \\token-level contextual representations using contrastive learning}

\author{Anna Mosolova$^{1,2}$, {\bf Marie Candito$^1$} , {\bf Carlos Ramisch$^2$} \\
         $^1$Université Paris Cité, CNRS, LLF, Paris, France \\ 
         $^2$Aix Marseille Univ, CNRS, LIS, Marseille, France \\ \texttt{first.last@u-paris.fr}, \texttt{first.last@lis-lab.fr}}

\begin{document}
\maketitle
\begin{abstract}

While static word embeddings are blind to context, for lexical semantics tasks context is rather too present in contextual word embeddings, vectors of same-meaning occurrences being too different \cite{ethayarajh-2019-contextual}.
Fine-tuning pre-trained language models (PLMs) using contrastive learning was proposed, leveraging automatically self-augmented examples \cite{liu-etal-2021-mirrorwic}. In this paper, we investigate how to inject a lexicon as an alternative source of supervision, using the English Wiktionary. We also test how dimensionality reduction impacts the resulting contextual word embeddings. We evaluate our approach on the Word-In-Context (WiC) task, in the unsupervised setting (not using the training set). We achieve new SoTA result on the original WiC test set. We also propose two new WiC test sets for which we show that our fine-tuning method achieves substantial improvements.  We also observe improvements, although modest, for the semantic frame induction task. Even if we experimented on English to allow comparison with related work, our method is adaptable to the many languages for which large Wiktionaries exist. 

\end{abstract}

\section{Introduction}
Pretrained language models (PLMs) have brought great advances in most NLP tasks. As far as word embeddings are concerned, though,
we have moved from one extreme to the other, namely
from static word embeddings providing a single representation for a given form, no matter how ambiguous it is, to contextual token embeddings providing one representation per occurrence. For lexical level tasks, while it is desirable that token-level vectors of the same word sense are close in the semantic space, this is not the case for the majority of PLMs \cite{ethayarajh-2019-contextual}.

In this paper, we address the tuning of token-level contextual representations to better target the lexical sense instantiated by a given token.
We use the contrastive learning (CL), which proved efficient for getting sentence embeddings that better capture sentence-level similarity \citep{reimers-2019-sentence-bert,gao-etal-2021-simcse,chuang-etal-2022-diffcse,fang_cert_2020} and for getting better token-level embeddings \citep{liu-etal-2021-mirrorwic,su-etal-2022-tacl}.
These approaches use self-supervised CL, with positive examples created by pairing an original sentence and an automatically modified version of it. 

In this paper, we rather investigate how to leverage hand-crafted lexicons. Although these are not always perfectly tailored to NLP tasks, due to coverage and granularity mismatches with the task or domain at hand, they do contain an enormous amount of lexical information that is a pity not to make use of. To do so, we use CL on the example sentences of the English Wiktionary, a crowd-sourced lexicon. We will show the approach is beneficial for both the Word-in-Context (WiC) task (intrinsic evaluation), and for the frame induction task (extrinsic evaluation). Crucially, although we experiment on English to allow comparison with related work, our method is adaptable to a large number of languages for which large Wiktionaries exist. 

We also investigate whether reducing dimensions can provide better-suited token-level contextual embeddings. 

In the following, we describe related work (\S~\ref{sec-related-work}), and how we adapted the CL loss to Wiktionary examples (\S~\ref{sec-lm-finetuning}). We present our language model fine-tuning experiments, along with an evaluation on the Word-in-Context task (\S~\ref{sec-experiments}). We test whether our fine-tuned token embeddings can help cluster verbal occurrences into semantic frames (\S~\ref{sec-frame-induction}). 

\section{Related Work}
\label{sec-related-work}

Within the deep metric learning paradigm, contrastive learning (CL) became increasingly popular 
in computer vision and in NLP \citep{kaya2019deep}. It consists in modifying the representation space so that similar objects (positive examples) are brought closer while dissimilar objects are pushed away from each other. \citet{Hadsell_Dimensionality} proposed one of the first contrastive loss functions, for binary positive examples.
CL methods are either supervised or self-supervised. While the former rely on labeled data, the latter employ automatic modifications of objects to produce binary positive pairs (self-augmentation). Since there can be more than two examples of the same class, \citet{khosla2020supervised} adapt the contrastive loss to handle ``multiple-positive'' examples for computer vision.

In NLP, CL is primarily used to improve sentence representations, better capturing sentence similarity, mainly in the self-supervised paradigm. Self-augmentation techniques include back translation \citep{fang_cert_2020}, text corruption \citep{liu-etal-2021-fast}, or PLM's dropout to produce slightly different embeddings per encoding run \citep{gao-etal-2021-simcse,chuang-etal-2022-diffcse}. \citet{zhuo-etal-2023-whitenedcse} combine whitening and CL to fine-tune sentence representations by PLMs.
Supervised CL is much less common. We can only cite \citet{gunel_supervised_2021} who use it for fine-tuning a PLM while learning a downstream sentiment-analysis classifier.




In contrast to sentence embeddings, fewer works focus on token-level PLM representations. \citet{liu-etal-2021-fast,liu-etal-2021-mirrorwic} fine-tune contextual embeddings using self-supervised CL, creating positive pairs with dropout and random masking of context tokens. \citet{su-etal-2022-tacl} use CL to favor more isotropic token-level representations. They train a student BERT model on the masked language modeling task with a help of a frozen teacher BERT model: CL aims at increasing the similarity of student and teacher token representations. 

Apart from CL, there was also work in enhancing BERT with senses during pre-training. For example, \citet{levine-etal-2020-sensebert} add supersense prediction for every masked word as pre-training objective. 

Finally, since we heavily rely on similarities of contextual embeddings, we mention studies reporting the particularities of such spaces. \citet{timkey-van-schijndel-2021-bark} show that very few dimensions dominate the cosine similarity 
and propose postprocessing methods to smooth this effect. \citet{zhou-etal-2022-problems} identified and \citet{wannasuphoprasit-etal-2023-solving} tried to solve the problem of underestimated cosine similarity for high-frequency words. 

Our goal is to obtain token-level contextual representations more aware of lexical semantics, by injecting lexicon-based information using CL. We show that this injection is beneficial for the closely related WiC task, and, to some extent, for the more downstream task of frame induction. 

\section{CL for lexical sense examples}
\label{sec-lm-finetuning}
Our method fine-tunes the token-level contextual representations of a PLM using supervised CL, taking the examples of a lexicon as supervision. More precisely, each example sentence in the lexicon is associated with a word sense and contains a target word occurrence used in this particular sense. 

We adapt the multiple-positive contrastive loss of \citet{khosla2020supervised} to the use of a lexicon as labeled data.\footnote{\citet{khosla2020supervised} test two formulations, varying in 
the precedence of log and summation 
over the same-class examples. They empirically show the superiority of applying log first. \citet{gunel_supervised_2021} also adopt this formulation.} Let $E(l)$ be the set of example sentences for  lemma $l$. For an example $j \in E(l)$, let $S(j)$ be the subset of $E(l)$ of examples concerning the same word sense as $j$, except for $j$ itself.
For every lemma $l$, we create a single batch, and we define a loss summing over the set $E(l)$ of all examples of $l$:
\begin{equation*}
    \mathcal{L}(l) = \sum_{j \in E(l)}  \frac{-1}{|S(j)|}\sum_{j' \in S(j)}\log\frac{e^{s(j, j')/\tau}}{\sum\limits_{k \in E(l) \setminus j}e^{s(j, k)/\tau}} 
\end{equation*}

 with $E(l)\setminus j$ being $E(l)$ except $j$. We write $s(m,n)$ for the similarity between the embeddings of the target tokens in examples $m$ and $n$ ($s$ can be any vector similarity function), and $\tau$ is a scalar temperature hyperparameter. 

In order to cope with known flaws of cosine similarity for high-dimensional spaces, we also experiment with a simple PCA reduction of the PLM embeddings, with or without whitening.

\section{PLM fine-tuning experiments}
\label{sec-experiments}

\paragraph{Training dataset}
\label{sec-training-dataset}

More precisely, our training data includes the examples for all verbs having from 1 to 10 senses, except verbs having a single sense with a single example, and multiword verbs. In total, we obtained a dataset of 13,118 verbs having in total 26,398 senses, with a total of 68,271 examples. Mean number of examples per sense is 2.59 {(std. dev. is 5.41)}. Mean number of senses per verb is 2.01 {(std. dev. is 1.54)}. Mean number of examples per verb is 5.21 {(std. dev. is 12.68)}. Each example concerns a target verb occurrence. For hyperparameter tuning and evaluation, we split the dataset into 95/5/5\% for training, development and test sets, ensuring that verb lemmas do not overlap between the three sets. 

\paragraph{Training details}
We report experiments using the bert-base-uncased model 
\citep{devlin-etal-2019-bert}.\footnote{\citet{pilehvar-camacho-collados-2019-wic} report BERT as the best-performing model in the unsupervised setting for the WiC task (\S~\ref{intrinsic_evaluation}). We used the -base instead of -large model to reduce the computational cost.} For the similarity metric (the $s$ function), 
we settled for cosine after a few experiments with various similarity metrics (euclidean distance, dot product).

The training procedure iterates for $E$ epochs, each epoch looping over shuffled training batches (one batch per lemma). We limited the batches' size by randomly selecting at most 64 examples per lemma 
($\max(|E(l)|)=64$). 
For a given batch, each example sentence $j$ is encoded using the current version of the PLM. The similarities $s(m,n)$ are computed by extracting the embedding, at the last layer, of the target tokens in $m$ and in $n$.\footnote{Sub-word token embeddings are averaged per word.}

\paragraph{Intrinsic evaluation: Word-in-Context (WiC)} 
\label{intrinsic_evaluation}


 is a binary classification task taking as input a pair of sentences containing the same target lexical unit, and predicting whether this target unit is used with the same meaning or not \cite{pilehvar-camacho-collados-2019-wic}. We use this task both to tune our CL method and to evaluate its benefits. We stress that since our objective is to evaluate contextual embeddings, we only consider the unsupervised scenario of the WiC task. Hence, we do not use the training WiC data at all.

For our hyperparameter tuning and evaluation, we use three kinds of WiC data {\bf (i) WiktWiC} is the data closest to our training data, namely the dev and test Wiktionary example set mentioned in \S~\ref{sec-training-dataset}, {\bf (ii) OrigWiC} are the original dev and test sets of the WiC task dataset\footnote{The original WiC dataset contains examples from VerbNet, WordNet and Wiktionary \cite{pilehvar-camacho-collados-2019-wic}. We deleted from all our Wiktionary dataset (train, dev, and test) \emph{all} examples in OrigWiC.} and {\bf (iii) FramenetWiC}, containing FrameNet 1.7 example pairs for the same verb, annotated with the same or different frames. Statistics for these datasets are provided in Appendix \ref{app-stats-wic}, Table \ref{tab:wic_datasets}. Each dataset is balanced for positive and negative pairs, hence the default metric is macro-averaged accuracy.

We perform the WiC task by applying a threshold on the cosine similarity between the target token embeddings (at the last layer) for the input sentences. Thus, we evaluate the impact of fine-tuning on the embeddings, without the influence of any additional classifier. The threshold is tuned with step size 0.02 on the development sets.

\paragraph{Hyperparameter tuning}
\label{hyperparameter_tuning}

To tune the hyperparameters, we used as a criterion the WiC accuracy, macro-averaged on the three development sets (Table~\ref{tab:wic_datasets}). The tested values and their results are provided in Appendix~\ref{app:hyperparam-details}, Table~\ref{tab:wic_evaluation_dev}. We chose the hyperparameter combination leading to the highest accuracy on average for the five runs, namely: learning rate = 5e-6, 2 epochs, temperature=0.5, PCA with whitening and 100 components. 


\begin{table}
    \centering
    \begin{adjustbox}{max width=0.48\textwidth}
    \begin{tabular}{ccccc}
    \hline
    FT & PCA & Wikt & Frame & Orig \\
     &  & WiC & WiC & WiC \\
    \hline
    - & - & 55.9 & 67.3 & 65.4 \\
    \hline
    - & + & 59.6 & 72.4 & 68.4 \\
    + & - & 70.0\small($\pm$0.9) &  69.6\small($\pm$0.4)$^{\ref{fn:repeat}}$ & 69.6\small($\pm$0.6) \\
    + & + & \textbf{70.5}\small($\pm$0.8) & \textbf{73.1}\small($\pm$0.4) & \textbf{71.4}\small($\pm$0.2) \\
    
    \hline
    \multicolumn{2}{c}{MirrorWiC}& - & - & 69.6 \\
    \hline
    \end{tabular}
    \end{adjustbox}
    \caption{Results on WiC test sets. \textbf{FT}: with or without fine-tuning.
    \textbf{PCA}: with or without PCA dimensionality reduction (100 components, with whitening). FT=+ rows are averages of 5 runs (std.~dev.~in parentheses).}
    \label{tab:wic_evaluation}
\end{table}

\paragraph{Unsupervised WiC results}
 As a baseline, we use the bert-base-uncased model, without applying PCA (first row of Table~\ref{tab:wic_evaluation}). The results are statistically significant\footnote{Except for the result of the fine-tuned model without PCA on the Frame WiC dataset, where the improvement was statistically significant on 3 runs out of 5.\label{fn:repeat}} in comparison with the baseline according to McNemar's test with $\alpha = 0.05$. We observe that our fine-tuning improves results for the three test sets. The best improvement is for the test set of the closest kind (WiktWiC), but improvements are also substantial for the two other test sets, which shows the method generalizes to other kinds of sense definitions, of varying granularity. We further observe that PCA is beneficial when applied to plain BERT embeddings, and the improvements add up when applying both fine-tuning and PCA.

\begin{table*}
    \centering
    \begin{adjustbox}{max width=\textwidth}
    \begin{tabular}
    {ccccccccc}
    \hline
    Model & Layer & $\alpha_{2}$ & \#pLU & \#C & Pu/iPu/PiF$_{1}$ & BcP/BcR/BcF$_{1}$ & Pu/iPu/PiF & BcP/BcR/BcF \\
    \hline
B & 11/2 & 0.6 & 1059 & 313 & 95.3/\textbf{99.6}/\textbf{96.8} & 94.4/\textbf{99.5}/\textbf{96.0} & 65.0/\textbf{75.5}/69.8 & 56.3/\textbf{67.1}/61.3 \\
B+P & 10/2 & 0.5 & 1083 & 307 & 95.5/99.2/96.7 & 94.7/98.9/95.9 & 65.3/72.2/68.6 & 54.7/62.4/58.3 \\
B+FT & 11/2 & 0.1 & 1228 & 394 & \textbf{97.4}/96.3/96.3 & \textbf{96.7}/95.3/95.2 & 68.4/72.2/70.2 & 59.8/62.9/61.3 \\
B+FT+P & 11/2 & 0.2 & 1157 & 381 & 96.6/97.8/96.7 & 95.8/97.2/95.7 & \textbf{69.9}/73.6/\textbf{71.7} & \textbf{60.5}/63.9/\textbf{62.1} \\

    \hline
    \end{tabular}
    \end{adjustbox}
    \caption{Results on the frame induction test set of Y21. {\bf B}: bert-base-uncased, {\bf P}: with PCA (100 components, with whitening), {\bf FT}: with our fine-tuning. {\bf Layer x/y}: layer x used for 1st step, and y for 2nd step clustering. {\bf $\alpha_2$}: weight of the masked embedding for the 2nd step. {\bf \#pLU}: number of pseudo-lexical units after the 1st step, {\bf \#C}: number of clusters after the 2nd step. Clustering algorithms are X-means (1st step) and group-average (2nd step). Gold number of LUs is 1,188, actual number of frames is 393. FT=+ rows report averages of 5 runs. {\bf Pu/iPu/PiF$_1$}: purity, inverse purity, and Fscore for the first step. {\bf BcP/BcR/BcF$_1$}: B-cubed precision/recall/Fscore for the first step. {\bf Pu/iPu/PiF} and {\bf BcP/BcR/BcF}: same but for the 2nd step.}
    \label{tab:framenet_test}
\end{table*}

We also compare our results on the OrigWiC dataset to MirrorWiC  \citep{liu-etal-2021-mirrorwic}, which leverages self-supervised CL to improve the last 4 layers of the token-level PLM embeddings.
Our approach outperforms MirrorWiC
, which shows that supervision even from a crowd-sourced lexicon surpasses the use of self-augmented examples. 
To the best of our knowledge, 71.4\% is the new state-of-the-art on the OrigWiC test set in the unsupervised setting, and it even surpasses some supervised settings that use the OrigWiC training set (see  \citet{loureiro2022lmms}).


\section{Extrinsic evaluation : frame induction}

\label{sec-frame-induction}
We now turn to evaluating our fine-tuning approach on semantic frame induction. Compared to word sense induction, frame induction seeks to identify semantic classes (or frames) that may group senses of different lemmas. It is thus a challenging task for token embeddings. We reuse the dataset of \citet{yamada2021semantic} (hereafter {\bf Y21}), extracted from the lexicographic part of Framenet 1.7.

We reproduce the approach of Y21 with minor modifications. It takes as input a set of words, each in the context of a sentence. Occurrences of the same lemma are clustered first, and the resulting clusters (called pseudo-lexical units) are then averaged and further clustered to form frames. To represent the target words to cluster, Y21 use a weighted average of two token embeddings obtained after applying a PLM on the original sentence, with and without masking the target word. 
We describe our minor modifications and hyperparameter tuning on Y21's dev set in Appendix~\ref{appendix-frame-induction-tuning}. 

We select the best hyperparameter combination (using the F-B-Cubed metric of the second clustering step) for each of the four types of embeddings: with and without CL fine-tuning, and with and without PCA. 
Results on the test set are provided in Table~\ref{tab:framenet_test}, for the four systems\footnote{For plain BERT, we were unable to 
reproduce Y21's results (PiF=73.0\%,  BcF=64.4\%), despite extensive tests. This might be due to hyperparameters left implicit in their description. We could not obtain answers from the authors.} (results on the dev set are in Table~\ref{tab:framenet_dev}, Appendix~\ref{sec:appendix_framenet_dev}). 
We did not perform the statistical significance test for this task, as it would require using bootstrapping which is extremely costly given that a new clustering must be created for each resampled pseudo-test set. 
For the first step, fine-tuning improves Purity and B-Cubed Precision, which means that clusters identified with the fine-tuned model contain less noise. However, items from the same frame tend to be divided into several clusters. With the two-step algorithm, such errors are recoverable, as the additional clusters can be merged during the second step, whereas over-merging cannot be undone by the second step. 

For frame induction (second step), while for the dev set our CL fine-tuning is clearly beneficial (+5.1 points for BcF), the increment on the test set is more modest and is only obtained with PCA ($62.1$ compared to $61.3$). The utility of CL fine-tuning for this task is thus limited, but with PCA it provides shorter embeddings, reducing computational cost for downstream tasks. 

We also notice that the best layers are high layers for the first step, but low layers for the second step. Moreover, after fine-tuning, the tuned $\alpha_2$ is close to 0, suggesting that flaws of the original unmasked token representations that were fixed when combining with the masked embeddings, were smoothed away during the fine-tuning step.

\section{Conclusions}
We presented a new approach for fine-tuning token-level representations of PLMs, using contrastive learning with examples from the English Wiktionary, a crowd-sourced lexicon. We show its effectiveness on the Word-in-Context task: we establish the new SoTA on the WiC test set, in the unsupervised setting (not using the WiC training set), and we also obtain substantial gains on two new WiC test sets, with different sense inventories. We also report improvements, though more modest, on the downstream task of semantic frame induction. Although we experimented on English, our method is adaptable to the many languages for which large Wiktionaries exist and provides a simple way to obtain token-level embeddings more adapted for lexical semantic tasks. A promising continuation of this work is to create positive examples using Wiktionary example sentences for distinct lemmas. 

\section{Limitations}

This paper proposes a new approach for fine-tuning token-level representations of PLMs. Our study is based on fine-tuning a single bert-base-uncased model. We believe that fine-tuning of its large version or other PLMs should also be studied to prove the generalisability of the method. Additionally, we conduct our experiments only using datasets in the English language. Our assumption of its applicability to other languages must also be tested in future work. As for the training dataset, we use only verbal lemmas for its constriction. However, it should be verified whether using lemmas of all parts of speech improves or worsens the fine-tuning results. 

We show the limited utility of CL fine-tuning for the frame induction task compared to the improvements achieved on the WiC datasets. We used only a single extrinsic task due to space limitations. Other lexical level tasks, such as word sense induction, can also be easily applied to investigate further abilities of the new representations (e.g. Task 14 of SemEval-2010 \cite{manandhar-etal-2010-semeval}).

\section*{Acknowledgements}
We thank the reviewers for their valuable feedback on our work. 

This work has been funded by the French Agence Nationale pour la Recherche, through the SELEXINI project (ANR-21-CE23-0033-01).

\bibliography{anthology,custom}
\bibliographystyle{acl_natbib}

\appendix

\section{Appendix}
\label{sec:appendix}

\subsection{Statistics for the three Word-in-Context datasets}
\label{app-stats-wic}
We provide the statistics for the three WiC datasets in table \ref{tab:wic_datasets}. We introduce 2 datasets: Wikt-WiC, which is a derivative of the Wiktionary DBnary dataset distributed under the Creative Commons Attribution-ShareAlike 3.0 license, and Framenet-Wic, which is created from the Framenet 1.7 examples \citep{fillmore2010frames}\footnote{\url{http://framenet.icsi.berkeley.edu/}} shared under the Creative Commons Attribution-Only license. We also reuse the original WiC dataset distributed under the Creative Commons Attribution-NonCommercial 4.0 license.

\begin{table}[h]
    \centering
    \begin{tabular}{ccc}
        \hline
        Dataset & Dev & Test \\
        \hline
        Orig-WiC & 638 & 1400 \\
        Wikt-WiC & 1200 & 1200 \\
        Framenet-WiC & 1800 & 1700 \\
        \hline
    \end{tabular}
    \caption{Statistics for three WiC evaluation datasets.}
    \label{tab:wic_datasets}
\end{table}

\subsection{Hyperparameter tuning of BERT fine-tuning by contrastive learning with Wiktionary examples, on the development sets of the WiC task}
\label{app:hyperparam-details}

We tuned the following hyperparameters using grid search: learning rate (tested values: 5e-7, 1e-6, 5e-6, 1e-5, 3e-5, 5e-5), number of epochs (from 1 to 6), temperature\footnote{We did some preliminary tests with all values from 0 to 1 with the step 0.1, and we finally only tested values 0.3 and 0.5 for the grid search.}, whether to use PCA or not (with or without whitening and number of components (tested values: from 100 to 700 with the step 100). 

We made five runs for each hyperparameter combination to determine the variance of the results. \\

Table \ref{tab:wic_evaluation_dev} shows the top 10 hyperparameter combinations of the bert-base-uncased CL fine-tuning. Additionally, we report results without fine-tuning as a baseline and MirrorWiC results on the development set (results from \cite{liu-etal-2021-mirrorwic}). 

The average training time of the bert-base-uncased model\footnote{\url{https://huggingface.co/bert-base-uncased}} (110M parameters) for one epoch is 30 minutes on one 4Gb GPU. For the fine-tuning, we used Transformers and SentenceTransformers libraries \citep{reimers-2019-sentence-bert}. We also use PCA implementation from the scikit-learn library \citep{scikit-learn}.
\begin{table*}
    \centering
    \begin{adjustbox}{max width=\textwidth}
    \begin{tabular}{ccccccccc}
        \hline
         LR & E & $\tau$ & N comp. & Whitening & Macro-Accuracy & Orig-WiC & Framenet-WiC & Wikt-WiC \\
         \hline
        \multicolumn{3}{c}{bert-base-uncased} & - & - & 65.6 & 67.9 & 70.9 & 58.0 \\
        \multicolumn{3}{c}{bert-base-uncased} & 100 & True & 67.5 & 69.6 & 73.9 & 58.9  \\ 
         \hline
        5e-6 & 2 & 0.5 & 100 & True & \textbf{71.4}\small($\pm$0.1) & 73.5\small($\pm$0.5) & 76.0\small($\pm$0.2) & 64.8\small($\pm$0.5) \\
        5e-6 & 3 & 0.5 & 100 & True & 71.4\small($\pm$0.2) & 73.7\small($\pm$0.4) & 75.8\small($\pm$0.2) & 64.8\small($\pm$0.3) \\
        5e-6 & 3 & 0.5 & 300 & True & 71.4\small($\pm$0.4) & 72.0\small($\pm$0.7) & 77.6\small($\pm$0.4) & 64.4\small($\pm$0.4) \\
        5e-6 & 2 & 0.5 & 300 & False & 71.3\small($\pm$0.2) & \textbf{73.9}\small($\pm$0.4) & 74.6\small($\pm$0.2) & 65.3\small($\pm$0.4) \\
        5e-6 & 2 & 0.5 & 300 & True & 71.3\small($\pm$0.4) & 71.9\small($\pm$0.6) & \textbf{77.8}\small($\pm$0.3) & 64.1\small($\pm$0.6) \\
        5e-6 & 3 & 0.5 & 400 & True & 71.2\small($\pm$0.4) & 72.0\small($\pm$0.8) & 77.5\small($\pm$0.4) & 64.1\small($\pm$0.5) \\
        5e-6 & 3 & 0.5 & 200 & True & 71.2\small($\pm$0.2) & 72.6\small($\pm$0.5) & 76.7\small($\pm$0.2) & 64.3\small($\pm$0.4) \\
        5e-6 & 2 & 0.5 & 200 & False & 71.2\small($\pm$0.3) & 73.5\small($\pm$0.5) & 74.6\small($\pm$0.3) & \textbf{65.4}\small($\pm$0.3) \\
        5e-6 & 1 & 0.5 & 100 & True & 71.2\small($\pm$0.1) & 72.8\small($\pm$0.4) & 75.8\small($\pm$0.2) & 64.9\small($\pm$0.4) \\
        5e-6 & 2 & 0.5 & 400 & False & 71.1\small($\pm$0.3) & 73.6\small($\pm$0.5) & 74.5\small($\pm$0.2) & 65.2\small($\pm$0.4) \\
        \hline
        \multicolumn{3}{c}{MirrorWiC} & - & - & - & 71.9 & - & - \\
        \hline
    \end{tabular}
    \end{adjustbox}
    
    \caption{Results on the development set of the WiC task. \textbf{LR} is learning rate, \textbf{E} - number of epochs, \textbf{\textit{$\tau$}} - temperature parameter of the loss function, \textbf{N comp.} - number of components for PCA. Reported metric is accuracy, all values are an average of 5 runs (std. dev. in parentheses). First two lines are baseline results before fine-tuning.}
    \label{tab:wic_evaluation_dev}
\end{table*}

\subsection{Hyperparameter tuning for the frame induction experiments}
\label{appendix-frame-induction-tuning}

To represent the target words to cluster, Y21 use a weighted average of two token embeddings obtained after applying a PLM on the original sentence, with and without masking the target word. The used embedding for a target word is $\alpha \cdot \upsilon_{MASK} + (1-\alpha) \cdot \upsilon_{WORD}$. Y21 use $\alpha_1=1$ for the first step, and a tuned $\alpha_2$ for the second step. We also tune $\alpha_2$, but we rather use $\alpha_1=0$, namely a plain embedding of the target word, without any masking, as we observed no impact on the results. Another difference in our implementation is that we may use different BERT layers for the first and second clustering steps, while Y21 always use the same.
The hyperparameter tuning, on the development set, is the following:
\begin{itemize}
\item First step clustering algorithm:
    \begin{itemize}
    \item X-means with minimum and maximum number of clusters set to 1 and 15 respectively,
    \item Agglomerative clustering with group average linkage.
    \end{itemize}
\item Combination of BERT layers for first and second steps: out of the 144 layer combinations, we first selected the 10 best combinations using the bert-base-uncased model with $\alpha_2=0$ and checked only 10 best combinations with the rest of hyperparameters.
\item $\alpha_2$ : tested values from 0 to 1 with step 0.1.
\end{itemize}

We do not tune the following hyperparameters: 
\begin{itemize}
\item  Number of components for PCA is always 100 with whitening application (the best combination identified in the WiC tuning).
\item Algorithm for the second step: Agglomerative clustering with group average linkage (with termination criterion as defined by Y21). 
\end{itemize}

\subsection{Results of the frame induction task on the development set}
\label{sec:appendix_framenet_dev}

In the table \ref{tab:framenet_dev}, we present the results on the development set of the frame induction task. We can see the improvement of all results after fine-tuning and a small degradation of the results after the PCA application. However, the clustering time is shorter by 13\% when reduced embeddings are used (2 minutes vs 2.3 minutes). Also, we observe that $\alpha_2$ values are close to 0 after fine-tuning suggesting removing the masked embedding completely as the overall computation time will be reduced by 2 times without its application.

B-Cubed metrics are computed using f-b-cubed python library\footnote{\url{https://github.com/hhromic/python-bcubed}}, purity metrics are computed with scikit-learn \citep{scikit-learn}. 

\begin{table*}[t]
    \centering
    \begin{adjustbox}{max width=\textwidth}
    \begin{tabular}
    {ccccccccc}
    \hline
    Model & Layer & $\alpha_{2}$ & \#pLU & \#C & PiF$_{1}$ & BcF$_{1}$ & PiF & BcF \\
    \hline
    B & 11/2 & 0.6 & 266 & 141 & 96.6 & 95.9 & 76.3 & 70.3 \\
    B+P & 10/2 & 0.5 & 275 & 144 & 96.9 & 96.1 & 75.4 & 69.3 \\
    B+FT & 11/2 & 0.1 & 300 & 171 & \textbf{97.2} & \textbf{96.4} & \textbf{80.7} & \textbf{75.4} \\
    B+FT+P & 11/2 & 0.2 & 294 & 163 & 97.2 & 96.4 & 80.3 & 74.8 \\
    \hline
    \end{tabular}
    \end{adjustbox}
    \caption{Results on the frame induction development set. Model name corresponds to {\bf B} - bert-base-uncased, {\bf P} - application of PCA (reduction to 100 components with whitening), {\bf FT} - the fine-tuned version of the BERT model. The layer column indicates which BERT layer was used: left value stands for the first step clustering layer, right value is the second step clustering layer. First step clustering algorithm is always X-Means, second step - Group Average. $\alpha_2$ is the weight of the masked embedding for the second step. \#pLU is the number of pseudo-lexical units after the first step clustering, \#C is the number of clusters after the second step. Actual number of LUs is 300, actual number of frames is 169. Every FT=+ row reports an average of 5 runs.}
    \label{tab:framenet_dev}
\end{table*}

\end{document}